\documentclass[final]{cvpr}

\usepackage{times}
\usepackage{epsfig}
\usepackage{graphicx}
\usepackage{amsmath}
\usepackage{amssymb}
\usepackage{caption}
\usepackage{subcaption}
\usepackage{multirow}
\usepackage{array}
\usepackage{array, makecell}
\usepackage{booktabs}
\usepackage[dvipsnames]{xcolor}
\usepackage{pifont}
\newcommand{\cmark}{\ding{51}}%
\newcommand{\xmark}{\ding{55}}%

\usepackage[pagebackref=true,breaklinks=true,colorlinks,bookmarks=false]{hyperref}

\def\cvprPaperID{****} 

\def\cvprPaperID{****} 
\def\confYear{CVPR 2021}

\begin{document}

\title{Drawing Attention to Detail: Pose Alignment through Self-Attention for Fine-Grained Object Classification}  

\author{Salwa Al Khatib, Mohamed El Amine Boudjoghra, Jameel Hassan\\
{\tt\small\{mohamed.boudjoghra, jameel.hassan, salwa.khatib\}@mbzuai.ac.ae}}

\maketitle
\thispagestyle{empty}


\section{Introduction}

Intra-class variations in the open world lead to various challenges in classification tasks. To overcome these challenges, fine-grained classification was introduced, and many approaches were proposed. Some rely on locating and using distinguishable local parts within images to achieve invariance to viewpoint changes, intra-class differences, and local part deformations. Throughout the years, different models have been suggested to achieve high performance in this task. Works \cite{kong2017low, tan2019fine} adopted bilinear pooling which uses  pairs of feature vectors extracted from two networks to better represent class variations, while \cite{Lin2017CVPR, part2pose} aims to assist the global context with important fine-grained details.

\par One recent study \cite{part2pose} shows the importance of \textit{local parts localization} and \textit{parts alignment} in enhancing the robustness against pose variation, and improving the generalizability of the model when trained with optimal order of fine-grained local parts. In their architecture, the parts from a given image are optimally arranged by maximizing the similarity between a correlation matrix of a reference set of parts, and the one generated from the input parts following all possible permutations.   
\par Our approach, which is inspired by \cite{part2pose}, offers an end-to-end trainable attention-based parts alignment module, where we replace the graph-matching component used in it with a self-attention mechanism. The attention module is able to learn the optimal arrangement of parts while attending to each other, before contributing to the global loss. 
\textit{Code available at \href{https://github.com/salwaalkhatib/P2P-Net}{https://github.com/salwaalkhatib/P2P-Net}}

\section{Baseline Architecture}
\begin{figure*}[t]
     \centering
     \includegraphics[width=0.9\textwidth]{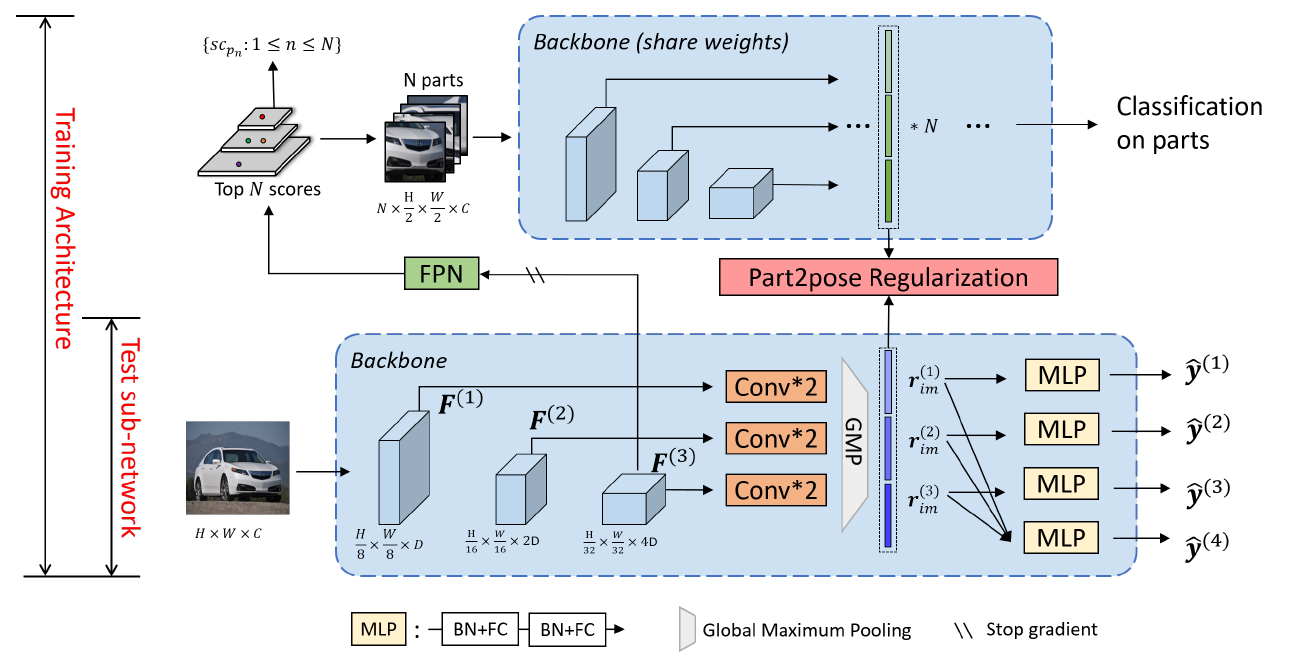}
     \caption{\centering Pipeline of P2P-Net \cite{part2pose}, the adopted baseline}
     \label{fig:pipeline}
\end{figure*}

We adopt the architecture of P2P-Net \cite{part2pose}, which is shown in Figure~\ref{fig:pipeline}, as the baseline model. In the following, we provide a brief description of the overall architecture and its key contributions.

\subsection{Overall architecture}
\begin{figure}
    \centering
    \includegraphics[width=0.6\linewidth]{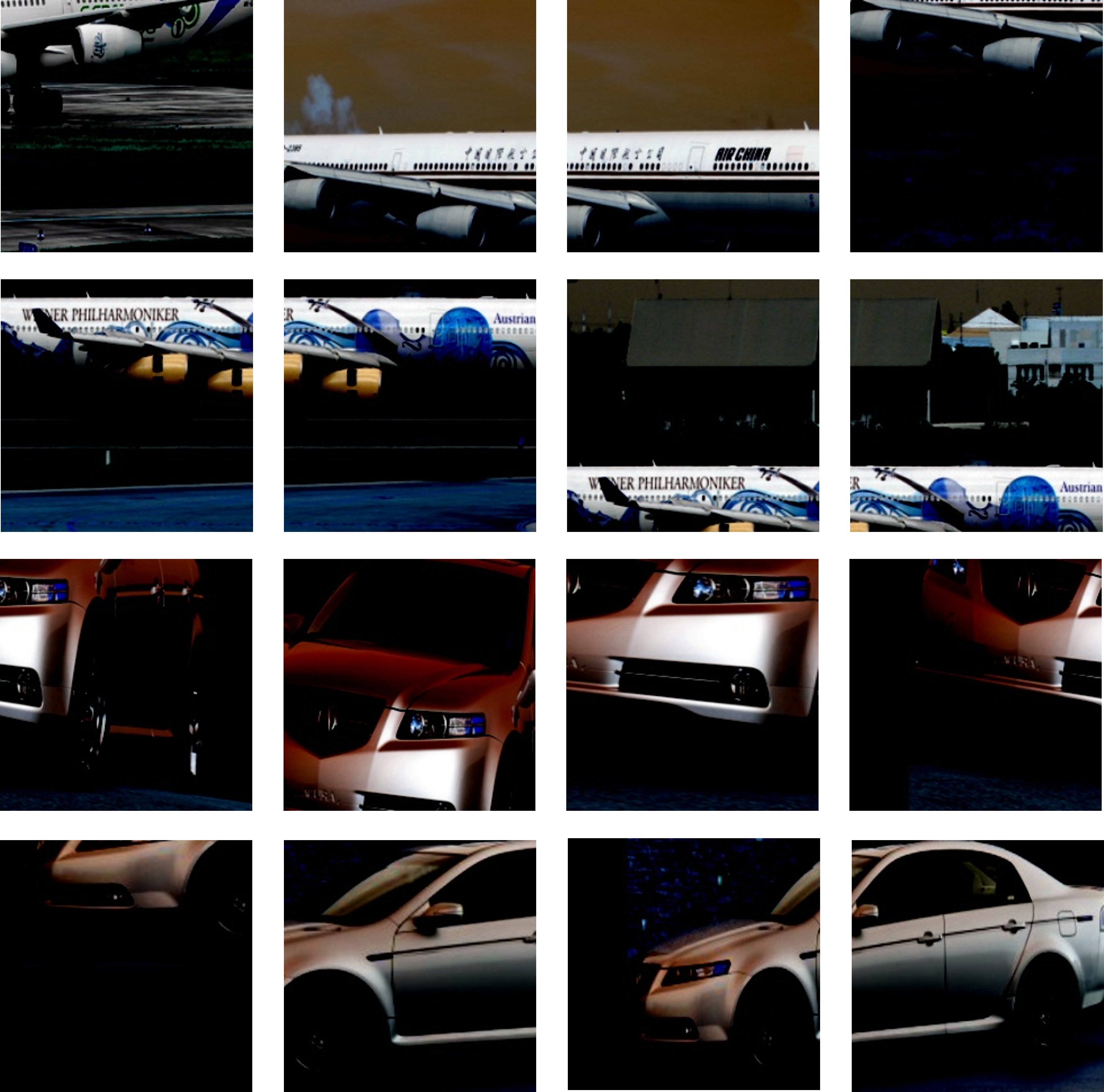}
    \caption{\centering Patches extracted by the feature pyramid network}
    \label{fig:parts}
\end{figure}

During training, the bottom global feature encoding backbone and the top local feature encoding backbone are used, while only the bottom network is activated for testing. The weights of the ResNet50 backbones are shared. Three feature maps ($F^{(1)}$, $F^{(2)}$, and $F^{(3)}$) at different depths are each fed into a separate set of convolutional blocks, followed by a global maximum pooling layer and a classification head. $F^{(3)}$, the model's last feature block, is fed into a feature pyramid network to generate $N$ patches that represent key parts of the salient object in the image. These patches are then fed to the top backbone that shares its weights with the bottom backbone. Examples of these patches can be seen in Figure~\ref{fig:parts}.

\subsection{Contrastive feature regularization}
To ensure that the model focuses on discriminative local details such as those learned by the top backbone, a contrastive loss is imposed on the model between $r_{im} = [r_{im}^{1}; r_{im}^{2};...;r_{im}^{S}]$ (the global image representations) and $r_{p_{n}} = [r_{p_{n}}^{1}; r_{p_{n}}^{2};...;r_{p_{n}}^{S}]$ (the local parts representations), as follows:
\begin{equation}
    L^{reg}=\sum_{s=1}^{S}\ell_{kl}\big(r_{im}^{(s)},\phi ([r_{p_{1}}^{s}; r_{p_{2}}^{s};...;r_{p_{N}}^{s}])\big)
\end{equation}

 where $S$ is the network depth, $\ell_{kl}$ is Kullback-Leibler divergence, and $\phi(.)$ is a 2-layer MLP used to approximate a unified representation of the same dimensionality as $r_{im}^{(s)}$. 
 
\subsection{Graph matching for part alignment}
\begin{figure}[htb]
    \centering
    \includegraphics[width=0.8\linewidth]{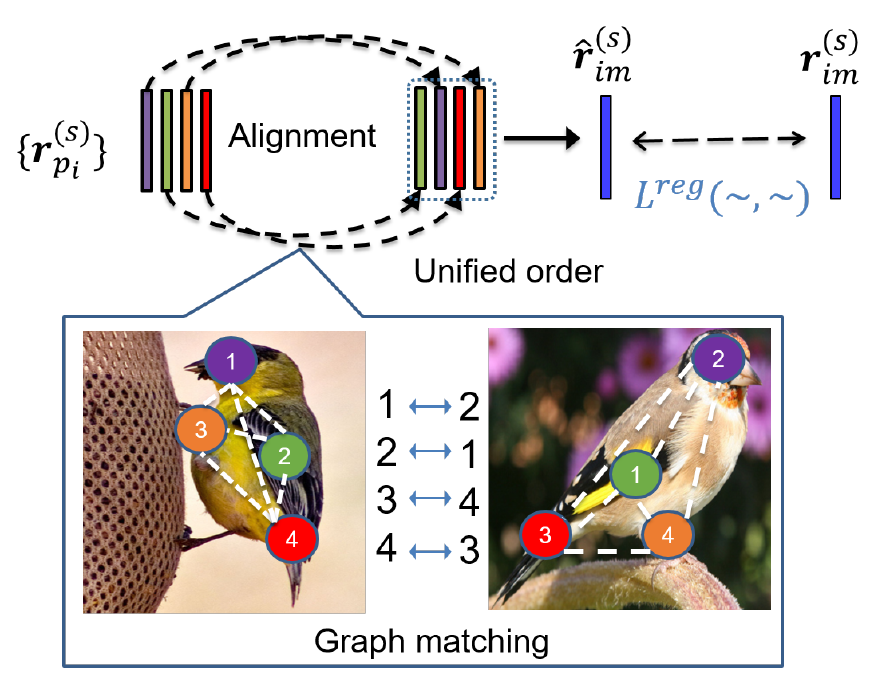}
    \caption{\centering Graph matching for pose alignment \cite{part2pose}. The feature maps later used for feature regularization are re-ordered through graph matching on the parts correlation matrix.}
    \label{fig:pose}
\end{figure}

The order of the detected top $N$ parts can not be expected to be consistent, and there is no straightforward way to re-order them since the exact semantic labels of the parts are unknown. This may cause an inconsistency problem for the later parts of the network since the features, e.g. head, body, and tail, are being fed to the subsequent layers in an arbitrary fashion. Thus, unsupervised graph matching is used to re-order the parts based on the correlation matrix between the $N$ parts across the different classes. This is illustrated in Figure~\ref{fig:pose}. This is done by means of a correlation matrix that is updated and maintained in memory throughout the training process. The re-sorted parts are then concatenated and mapped through $\phi(.)$ to obtain the representation that is fed into the KL divergence loss with the global features.

\section{Contributions}

The Part2Pose regularization block is a key component in the architecture focusing on fine-grained classification. This performs a feature regularization on representations between local parts and the global object. The inconsistency of the part order is handled by the graph-matching module by re-ordering them. We replace this block of graph matching component using a transformer block, passing the local part features through self-attention to obtain a refined representation of the local features. \par

\begin{figure}[htb]
    \centering
    \includegraphics[width=0.7\linewidth]{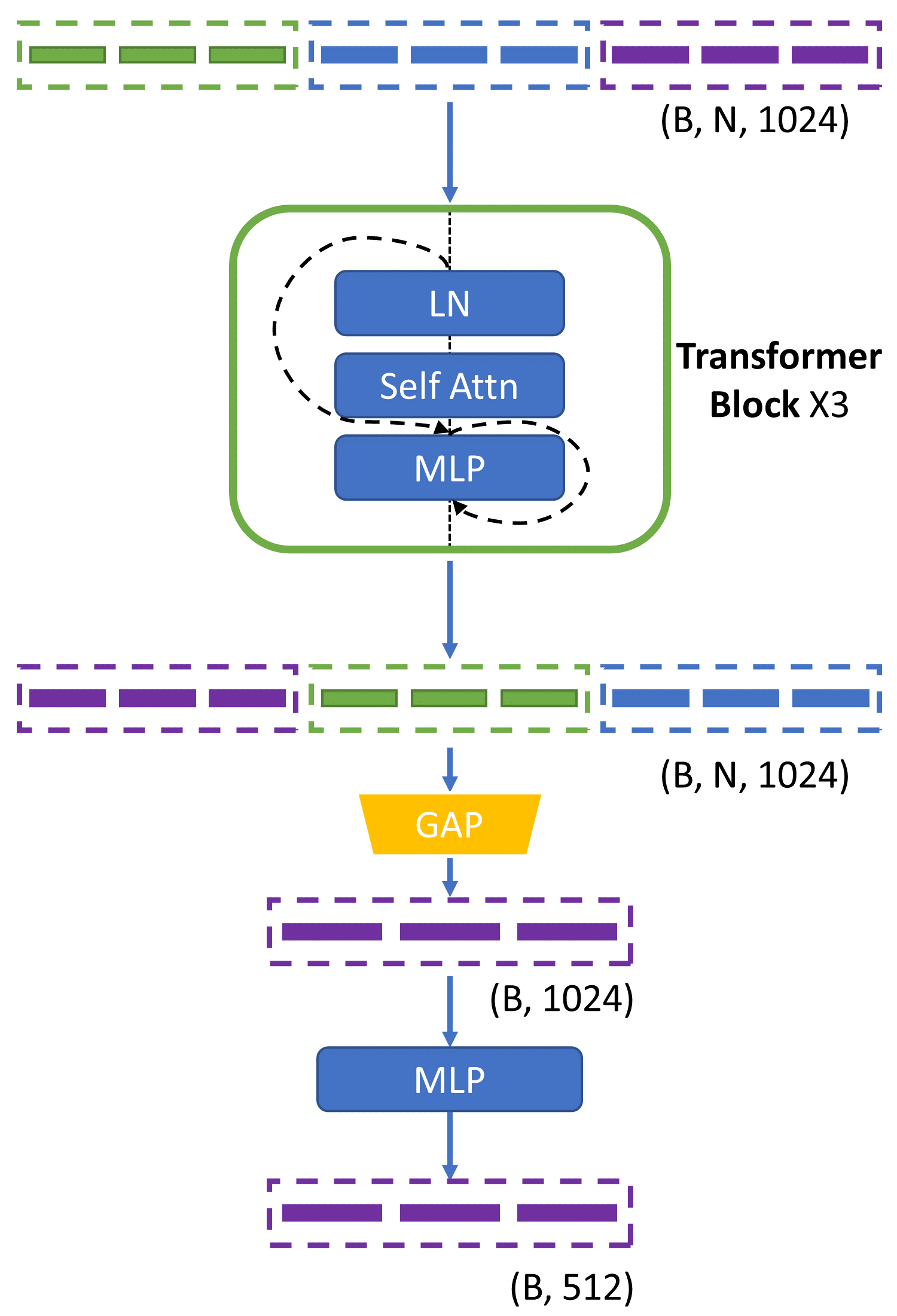}
    \caption{\centering Proposed self-attention mechanism in \texttt{Attn2Parts} to replace graph matching for pose alignment}
    \label{fig:attention}
\end{figure}

\begin{table*}[htb]
\centering
\caption{\centering Comparison of model performances}
\label{tab:results}

\begin{tabular}[\textwidth]{l c c c c}
    \specialrule{.2em}{.2em}{.2em}
    \textbf{Model}& \textbf{Color Jitter} & \textbf{AIR} & \textbf{AIR + CARS} & \textbf{FOOD101}\\
    \specialrule{.2em}{.2em}{.2em}
    Baseline & \xmark & 91.98\% & 93.3\% & 73.14\%\\
    \midrule
        Baseline & \cmark & 93.18\% & 93.87\% & -\\
    \midrule 
    \texttt{Attn2Parts} (3 layers) & \xmark & 93.01\% & 93.0\% & -\\
    \midrule
    \texttt{Attn2Parts} (1 layer) & \cmark & 93.6\% & 93.1\% & -\\      
    \midrule
    \texttt{Attn2Parts} (Cross Attn.) & \xmark & 88.84 & 89.99\% & -\\
    \midrule
    \texttt{Attn2Parts} (3 layers) & \cmark & \textbf{93.43\%} {\color{ForestGreen} (+1.45)} & \textbf{94.0\%} {\color{ForestGreen} (+0.7)} & 72.27\% {\color{Red} (-0.87) } \\
    \specialrule{.2em}{.2em}{.2em}
\end{tabular}
\end{table*}

We utilize three transformer blocks \cite{2017attention, ViT}, each consisting of a layer normalization, a self-attention layer, and an MLP as illustrated in Figure~\ref{fig:attention}. The output tokens are averaged using global average pooling and fed through an inverted bottleneck two-layer MLP to obtain the final representation of the local part features. Using self-attention, we remove the need for a memory bank to maintain the correlation matrix. We also get rid of the constraint to simply re-sort the parts as captured in the running correlation matrix, but rather learn better representative features through attention across parts. We also use color jitter as a data augmentation for the training process to improve the training. \par

We thus replace the graph matching component with self-attention and the mapping MLP function using a lighter MLP since a single representation of the local parts is fed in instead of all part features. 



\section{Results and Discussion}

The proposed architecture is compared with the baseline architecture on three different tasks. In the first task we evaluate the performance on the FGVC aircraft \cite{aircraft} dataset, the second is on an extended dataset consisting of both FGVC and Stanford Cars \cite{stanfordcar} and the last task is on the Food101 \cite{foodx} dataset. The models are trained for 100/150 epochs on all datasets. All hyperparameters are maintained the same for both the baseline and the modified proposed architecture (\texttt{Attn2Parts}). \par 

The models in comparison named \texttt{Attn2Parts}, replace the local-global alignment graph matching component with transformer layers. The proposed model uses three transformer blocks as in Figure~\ref{fig:attention}. We also experiment with a single transformer block and with cross attention, using global features as the queries and the local-part features as key-value inputs to the layer. These are identified with the naming in brackets as \textit{1 layer} and \textit{Cross Attn.}. We also report the performance with color jitter in our final model. \par 

The 3-layer transformer block improves the performance in FGVC aircraft dataset, but slightly drops in the extended dataset of aircraft and Stanford Cars. Using a 1 layer transformer block with color jitter shows a similar trend but better improvement. The use of cross-attention with global image features has the lowest accuracy. Since the pose alignment through local parts is used only during training, this needs to help the model learn its global features better through the KL-divergence loss. We hypothesize that the features learned through attention, by cross-attention between local and global features lose the granularity of local features, impairing learning. Thus, the final model which uses self-attention between local part features performs the best with an improvement of \textbf{+1.45} on FGVC aircraft dataset and \textbf{+0.7} on the extended aircraft and cars dataset. This supports our hypothesis earlier, where the tokens learned through attention are now representative of the local part features, which facilitates the global feature representation learning. On the Food101 dataset, the proposed model performance drops by \textbf{-0.87}. This could be due to the lack of notion of "parts" in food images, hence the local part features are redundant and do not contribute to the model learning in the classification task.

\section{Conclusion}
The proposed transformer-based alignment block showed clear superiority when classifying images with distinguishable local parts, where it surpassed the graph-based alignment with \textbf{+1.45\%} on FGVC aircraft dataset and \textbf{+0.7\%} on the extended dataset. On the other hand, the experiments on the Food101 dataset show the redundancy of attention when dealing with categories with redundant local parts.   


{\small

\bibliographystyle{IEEEtran}
\bibliography{main}
}


\end{document}